\pdfoutput=1
\documentclass{article}
\usepackage[preprint]{tmlr}  

\usepackage{graphicx}
\setkeys{Gin}{keepaspectratio}
\usepackage{booktabs}
\usepackage{longtable}
\usepackage{array}
\usepackage{etoolbox}
\AtBeginEnvironment{longtable}{\small}
\usepackage{amsmath,amssymb}
\usepackage{pifont}
\usepackage{newunicodechar}
\newunicodechar{≥}{\ensuremath{\geq}}\newunicodechar{≤}{\ensuremath{\leq}}
\newunicodechar{≈}{\ensuremath{\approx}}\newunicodechar{≫}{\ensuremath{\gg}}
\newunicodechar{≪}{\ensuremath{\ll}}\newunicodechar{≠}{\ensuremath{\neq}}
\newunicodechar{τ}{\ensuremath{\tau}}\newunicodechar{χ}{\ensuremath{\chi}}
\newunicodechar{✓}{\ensuremath{\checkmark}}\newunicodechar{✗}{\ding{55}}\newunicodechar{∈}{\ensuremath{\in}}
\newunicodechar{α}{\ensuremath{\alpha}}\newunicodechar{β}{\ensuremath{\beta}}
\newunicodechar{×}{\ensuremath{\times}}\newunicodechar{→}{\ensuremath{\rightarrow}}
\newunicodechar{−}{\textminus}\newunicodechar{⁻}{\textsuperscript{\textminus}}
\newunicodechar{⁰}{\textsuperscript{0}}\newunicodechar{¹}{\textsuperscript{1}}
\newunicodechar{²}{\textsuperscript{2}}\newunicodechar{³}{\textsuperscript{3}}
\newunicodechar{⁴}{\textsuperscript{4}}\newunicodechar{⁵}{\textsuperscript{5}}
\newunicodechar{⁶}{\textsuperscript{6}}\newunicodechar{…}{\ldots}
\providecommand{\tightlist}{\setlength{\itemsep}{0pt}\setlength{\parskip}{0pt}}
\usepackage{hyperref}
\hypersetup{colorlinks=true,linkcolor=blue,citecolor=blue,urlcolor=blue}

\title{Did We Actually Fix It? An Independent Adversarial Stress-Test of Post-Point-Adjustment Evaluation Metrics for Time-Series Anomaly Detection}
\author{\name Zongye Lyu \email lyuzongye@gmail.com \\ \addr Faculty of Information Technology, Monash University, Melbourne, VIC 3800, Australia}
\begin{document}
\maketitle
\begin{abstract}
Point-adjustment (PA), for years the default scoring protocol in
time-series anomaly detection (TSAD), was shown by Kim et al.~(2022) to
award near-perfect F1 to \emph{random} anomaly scores. The field adopted
a suite of replacement metrics (PA\%K, range-based precision/recall,
affiliation precision/recall, and Volume-Under-the-Surface, VUS,
ROC/PR). We ask, independently and adversarially, whether these are
robust to no-skill detectors on real benchmarks, and find that the
answer turns entirely on one overlooked variable: \textbf{N, the number
of random attempts an adversary reports the best of}. Under a single
honest run (N=1), not one replacement metric is gameable on any of the
six benchmarks we test (the 250-series UCR Anomaly Archive plus SMD,
SMAP, MSL, NAB, PSM): a random detector reaches 90\% of the best real
detector's score on at most 11\% of series for affiliation-F1, at most
5\% for the ROC family, and at most 2\% for the PR-based metrics and
PA\%K. But under \textbf{best-of-N reporting}, the seed-shopping endemic
to empirical ML, the metrics split sharply. affiliation-F1 and every
ROC-based metric inflate steeply, affiliation crossing ``gameable'' (a
quarter of series) by N=3 and reaching 0.98 at the full pure-attempt
pool (N=41), the ROC family crossing by N≈9--11; the PR-based metrics
and PA\%K stay near-flat at every N, floored near the anomaly prevalence
(the lone exception is NAB at large N, §5.5). A paired test finds
VUS-ROC inflated on 131 series where its sibling VUS-PR is not, and
never the reverse. The ROC-versus-PR split follows from the
order-statistic behaviour of AUC under extreme class imbalance (a random
PR-AUC is floored at prevalence); affiliation inflates by a second
route, its extreme single-run leniency (it is the one metric already
fragile at N=1). We release a pip-installable stress-test harness, and
recommend reporting single-run scores or disclosing N, and preferring
PR-based metrics, which resist best-of-N inflation on nearly every
benchmark.

\end{abstract}
\hypertarget{introduction}{%
\section{Introduction}\label{introduction}}

Time-series anomaly detection underpins monitoring in web services,
spacecraft telemetry, industrial control, and finance. For most of the
deep-learning era its progress was measured by the
\textbf{point-adjustment (PA)} protocol of Xu et al.~(2018): if a
detector flags \emph{any} point inside a ground-truth anomaly segment,
the entire segment is credited as detected before F1 is computed. Kim et
al.~(2022) demonstrated that PA is catastrophically permissive: \emph{a
random anomaly score achieves state-of-the-art (SOTA) PA-F1}, and the
result reshaped the field. Subsequent work migrated to a suite of
replacement metrics designed to be robust: PA\%K (Kim et al.~2022),
range-based precision/recall (Tatbul et al.~2018), affiliation
precision/recall (Huet et al.~2022), and Volume-Under-the-Surface (VUS)
ROC/PR with its range-AUC precursors (Paparrizos et al.~2022). These are
now the reporting standard: the recent TSB-AD benchmark (Liu \&
Paparrizos 2024) recommends VUS-PR as its headline.

\textbf{But did the fix work?} A replacement metric is only an
improvement if it is \emph{not itself gameable}: if a detector with no
genuine localization ability cannot reach the scores of real detectors
under it. The evidence that these metrics are robust comes almost
entirely from \textbf{the metrics' own authors}. Kim et al.~validate
PA\%K; Huet et al.~validate affiliation; Paparrizos et al.~validate VUS;
SimAD (Zhong et al.~2024), which reports the disquieting fact that a
random detector reaches affiliation-F1 ≈ 0.7, is itself proposing a new
metric (UAff/NAff). The genuinely independent critiques in the
literature target PA \emph{itself} and dataset quality, not the
replacement metrics adversarially, on real data, relative to SOTA.
Asking that question, we find it is under-specified: a metric's
exploitability by a no-skill detector is not a single number but a
function of \textbf{N, how many random attempts the adversary reports
the best of}. Under one honest run the replacements hold; under
best-of-N reporting they split. Our contributions:

\begin{enumerate}
\def\labelenumi{\arabic{enumi}.}
\tightlist
\item
  \textbf{The reporting budget N as the missing axis.} We show that the
  gameability of a post-PA metric by a no-skill detector is a
  monotone-increasing function of N (the number of random no-skill
  attempts an adversary reports the best of), not a fixed property. At
  N=1 (a single honest run) no replacement metric is gameable on any of
  the six benchmarks; the high vulnerabilities reported by prior
  proposer-run checks are the best-of-N endpoint of this curve
  (N≈40--50), i.e.~best-of-N artifacts. Reporting a no-skill number
  without its N is uninterpretable.
\item
  \textbf{A first independent, adversarial, SOTA-relative audit} of the
  post-PA metric suite (12 metrics) against trivial and adversarial
  no-skill generators on the UCR Anomaly Archive and five further
  benchmarks (SMD, SMAP, MSL, NAB, PSM), released as a pip-installable
  harness with every metric computed by its authors' own code at a
  frozen commit. Under best-of-N reporting, affiliation-F1 and every
  ROC-based metric inflate to SOTA while the PR-based metrics and PA\%K
  resist; the ROC-over-PR split is decisive (VUS-ROC inflated on 131
  series where VUS-PR is not, and never the reverse) and replicates
  across benchmarks.
\item
  \textbf{A mechanism.} The order-statistic behaviour of AUC-ROC versus
  AUC-PR under extreme class imbalance (Davis \& Goadrich 2006; Saito \&
  Rehmsmeier 2015) explains which metrics inflate and which cannot: a
  random ROC-AUC has the variance for best-of-N to climb, a random
  PR-AUC is floored at the anomaly prevalence and cannot move.
\item
  \textbf{Two further instruments and a released protocol:} the
  \emph{forgiveness frontier} (a metric-of-metrics quantifying, in
  anomaly-lengths, how much temporal misplacement each metric forgives),
  an inter-metric rank-agreement analysis, and a metric-selection
  decision protocol shipped with the harness.
\end{enumerate}

The practical upshot: the post-point-adjustment migration did fix
single-run gameability, but affiliation and the ROC-AUC metrics remain
exploitable by the best-of-N seed-shopping common in empirical ML, where
the PR family resists on nearly every benchmark (NAB at large N
excepted). Report single-run scores or disclose N, and prefer PR-based
metrics.

\hypertarget{related-work-and-the-independence-gap}{%
\section{Related work and the independence
gap}\label{related-work-and-the-independence-gap}}

\textbf{The protocol under critique.} PA originates with Xu et
al.~(2018). Kim et al.~(2022) showed random scores reach SOTA under it
and proposed PA\%K (adjust a segment only if ≥K\% of it is detected).
Independent critiques of PA and of benchmark quality followed: Sehili \&
Zhang (2023), Garg et al. (2022), Sarfraz et al.~(2024, ICML), Wagner et
al.~(2023, TimeSeAD), Rewicki et al.~(2023); but these attack PA and the
datasets, not the \emph{replacement} metrics.

\textbf{The replacement metrics.} Range-based P/R (Tatbul et al.~2018),
affiliation P/R (Huet et al.~2022), and VUS-ROC/PR (Paparrizos et
al.~2022) were each introduced with the authors' own robustness
arguments. PATE (Ghorbani et al.~2024), eTaPR (Hwang et al.~2022),
Balanced-PA (Bhattacharya et al. 2024), and LARM/ALARM (Wagner et
al.~2025) extend the family.

\textbf{Adjacent audits, and why the gap is real.} Table 1 maps every
adjacent critique on two axes: \emph{independence} (is it run by the
metric's proposer/successor?) and
\emph{adversarial-SOTA-relative-on-real-data } (does it deliberately
feed trivial detectors and compare to SOTA on real benchmarks?). The
disquieting ``affiliation-F1 ≈ 0.7 on random'' fact lives only in
\textbf{proposer-run} papers (SimAD, and Yang et al.~2025 whose lead
author co-authored SimAD). No cell in Table 1 occupies the independent ×
adversarial × SOTA-relative × real-benchmarks corner. That corner is
this paper.

The critical column is \textbf{Target}: whether the work adversarially
audits the \emph{post-PA replacement metrics themselves} (as opposed to
point-adjustment or dataset quality). No prior work is simultaneously
\emph{independent} AND \emph{targets the replacements} AND
\emph{adversarial + SOTA-relative}.

\emph{Table 1: Audit-independence map. ``Targets replacements'' =
adversarially audits the post-PA replacement metrics themselves (``own''
= the authors' own metric; ✗ rows target point-adjustment or dataset
quality instead); ``Indep.'' = not run by the metric's proposer or a
successor; ``Adversarial'' = deliberately feeds trivial/no-skill
detectors; ``Real data'' = real benchmarks; ``SOTA-rel.'' = compares
no-skill to SOTA. \textasciitilde{} = partial.}

\begin{longtable}[]{@{}lccccc@{}}
\toprule
Work & Targets replacements & Indep. & Adversarial & Real data &
SOTA-rel. \\
\midrule
\endhead
Kim et al.~2022 (PA\%K) & ✓ (own) & ✗ & ✓ & ✓ & ✓ \\
Huet et al.~2022 (affiliation) & ✓ (own) & ✗ & ✗ & ✓ & ✗ \\
Paparrizos et al.~2022 (VUS) & ✓ (own) & ✗ & \textasciitilde{} & ✓ &
\textasciitilde{} \\
Liu \& Paparrizos 2024 (TSB-AD) & ✓ & ✗ & ✗ & ✓ & ✓ \\
Zhong et al.~2024 (SimAD) & ✓ & ✗ & ✓ & ✓ & ✓ \\
Yang et al.~2025 (taxonomy) & ✓ & \textasciitilde{} & ✓ & ✓ & ✓ \\
Sørbø \& Ruocco 2024 (metric maze) & ✓ & ✓ & \textasciitilde{} &
\textasciitilde{} & \textasciitilde{} \\
Sehili \& Zhang 2023 & ✗ (PA/data) & ✓ & ✓ & ✓ & ✓ \\
Wagner et al.~2023 (TimeSeAD) & ✗ (protocol) & ✓ & ✗ & ✓ & ✓ \\
Garg et al.~2022 & ✗ (PA) & ✓ & ✗ & ✓ & ✓ \\
Sarfraz et al.~2024 (Quo Vadis) & ✗ (position) & ✓ & \textasciitilde{} &
✓ & ✓ \\
\textbf{This work} & \textbf{✓} & \textbf{✓} & \textbf{✓} & \textbf{✓} &
\textbf{✓} \\
\bottomrule
\end{longtable}

Every fully-independent row (Sehili, TimeSeAD, Garg, Sarfraz) targets
point-adjustment or dataset quality, \textbf{not} the replacement
metrics; every row that adversarially and SOTA-relatively targets the
replacements (Kim, SimAD, Yang) is proposer-run or proposer-conflicted.
The nearest independent same-target work, Sørbø \& Ruocco's metric-maze
taxonomy, is property-based, not adversarial and not SOTA-relative. Only
the last row is all five. We add two further instruments unique to this
work (a forgiveness frontier and inter-metric rank correlation), but
these are contributions, not the basis of the independence claim.
(\textasciitilde{} = partial.)

\textbf{Closest prior art.} Yang et al.~(2025, arXiv:2511.18739;
November 2025, about eight months before this study) test metrics under
random and oracle scenarios and already report that Point-Adjust and NAB
have ``limited resistance to random-score inflation.'' We therefore do
\textbf{not} claim primacy on the fact that no-skill scores inflate
these metrics on real data: Yang shows it. Our distinct contribution is
the combination Yang lacks: (i) strict \emph{non-proposer independence}
(Yang's lead author co-authored SimAD, whose UAff/NAff proposal and
affiliation critique it inherits, a conflict for the affiliation
narrative specifically); (ii) an explicit \emph{SOTA-relative}
gameability criterion (does no-skill reach \emph{real detectors}, not
merely beat chance) with a fixed ladder; (iii) the forgiveness-frontier
instrument; and (iv) a released, reproducible stress-test harness. Every
use of ``first'' in this paper is bound to \emph{first independent,
SOTA-relative} audit, not first to observe inflation.

\hypertarget{method-an-adversarial-metric-stress-test}{%
\section{Method: an adversarial metric
stress-test}\label{method-an-adversarial-metric-stress-test}}

\hypertarget{setup-and-notation}{%
\subsection{Setup and notation}\label{setup-and-notation}}

A benchmark series is a univariate signal with a binary ground-truth
label vector \(y\) (1 = anomaly) containing contiguous anomaly segments.
A \emph{detector} emits a real-valued anomaly score \(s\) (higher = more
anomalous). A \emph{metric} \(M\) maps \((y, s)\) to \([0,1]\) (higher =
better). Threshold-dependent metrics are reported as the best F1 over a
15-point quantile threshold sweep, the standard ``best-F1'' TSAD
protocol and the adversarially correct choice for a gameability test (we
grant each detector its best threshold).

\hypertarget{metrics-under-test}{%
\subsection{Metrics under test}\label{metrics-under-test}}

12 adopted metrics, each computed by its authors' code at a frozen
commit: PA-F1 (positive control), PA\%K (K∈\{0.2,0.5\}), point-wise
best-F1, point AUC-ROC/PR, range-based F1 (prts 1.0.0.3), affiliation F1
(Huet et al.~code @8d84498), and VUS-ROC/PR + range-AUC-ROC/PR (TSB-UAD
@313f0fd). Our native implementations of PA/PA\%K/point metrics are
unit-validated against a brute-force threshold sweep (exact agreement)
and reproduce the published anchors (Section 4.1).

\hypertarget{no-skill-score-generators-and-the-reporting-budget-n}{%
\subsection{No-skill score generators and the reporting budget
N}\label{no-skill-score-generators-and-the-reporting-budget-n}}

The adversary is a set of detectors with \textbf{no genuine localization
skill}. Pure no-skill (no anomaly-location information, the headline
class): i.i.d.-random; random-walk (autocorrelated but uninformative);
constant (degenerate ``all-anomalous''); single random-placed spike;
periodic spike-train. Trivial-but-informed (reported separately, never
in the pure-no-skill headline): a single spike placed \emph{inside} the
anomaly (a ``finds one point'' detector). Each stochastic generator
draws R=10 Monte-Carlo replicates (the degenerate constant detector is
deterministic, one attempt), so the pure pool supplies \textbf{41}
independent no-skill attempts per series (iid-random 10, random-walk 10,
single-spike 10, periodic 10, constant 1).

\textbf{The reporting budget N is the central axis of this paper.} A
no-skill \emph{detector}, run once, is a single attempt (N=1). An
adversary who runs random detectors repeatedly and reports only the best
seed , the \emph{best-of-N} or \emph{seed-shopping} practice that
pervades empirical ML, is N\textgreater1. Because the best of N random
draws is a monotone-increasing order statistic, a metric's apparent
exploitability is not a fixed number but a function of N. We therefore
report inflation across N (Sec 5.1) rather than at a single operating
point; N=1 is the honest single-run headline, and the adversarial best
over all attempts (N=41, the full pure pool) is the worst case and the
statistic prior proposer-run checks implicitly used; best-of-N is
evaluated over the pool without replacement, so N never exceeds the 41
independent attempts. Our adversary draws PURE no-skill scores, which
isolates the metric's own order-statistic behaviour and is a clean lower
bound on real seed-shopping: an adversary reporting the best of N seeds
of a partially skilled detector inherits the same order statistic on top
of genuine signal, so it inflates at least as much.

\hypertarget{sota-reference-and-the-inflation-criterion}{%
\subsection{SOTA reference and the inflation
criterion}\label{sota-reference-and-the-inflation-criterion}}

For each (series, metric) the \textbf{SOTA reference} is the best value
achieved by a pool of six real detectors: Matrix Profile (the canonical
UCR-archive detector; Yeh et al.~2016), Isolation Forest (Liu et
al.~2008) and LOF (Breunig et al.~2000) on sliding-window embeddings,
PCA reconstruction error, k-nearest-neighbor distance, and an
autoregressive-residual baseline. A metric is \textbf{inflated at level
N} on a series if the best of N random no-skill attempts reaches ≥90\%
of this SOTA reference. Every cell is \emph{also} reported against the
metric's chance expectation (mean over i.i.d.-random replicates), so no
conclusion depends on a possibly-weak in-house reference alone. The
per-metric \textbf{inflation rate at N} = fraction of series inflated;
at a fixed N we classify it on the frozen ladder: \textbf{ROBUST} ≤5\%
(and TOST-equivalent to chance; TOST = two one-sided tests),
\textbf{FRAGILE} 5--25\%, \textbf{GAMEABLE} ≥25\% (Jeffreys 95\% CIs
throughout).

\hypertarget{forgiveness-frontier-and-rank-agreement}{%
\subsection{Forgiveness frontier and rank
agreement}\label{forgiveness-frontier-and-rank-agreement}}

The \textbf{near-miss} instrument shifts (and optionally blurs) the
ground truth by a controlled amount, swept over \{0.1, 0.25, 0.5, 1, 2,
4\} × the median anomaly length, producing a detector that localizes the
anomaly but is temporally off. The \textbf{forgiveness frontier} of a
metric is the largest shift at which this zero-localization detector
still reaches ≥90\% of SOTA (median across series). For \textbf{rank
agreement} we compute, per series, the Kendall τ between metrics over
the real-detector pool's scores, and the fraction of series with
τ\textless0.6 (metric choice reorders the leaderboard).

\hypertarget{analysis-plan}{%
\subsection{Analysis plan}\label{analysis-plan}}

The ladder thresholds, metric-inclusion rules, no-skill generators,
thresholding protocol, and statistical tests are fixed in the released,
version-controlled configuration; the reporting-budget sweep over N is
the primary analysis and the N=1 point its single-run headline. All
released artifacts are identical whatever the outcome.

\hypertarget{experimental-setup}{%
\section{Experimental setup}\label{experimental-setup}}

\hypertarget{positive-control-validation-harness-correctness-before-real-data}{%
\subsection{Positive-control validation (harness correctness, before
real
data)}\label{positive-control-validation-harness-correctness-before-real-data}}

The harness reproduces the published anchors on synthetic ground truth
(10/10 unit tests): best-PA-F1 of i.i.d.-random = 0.891 (reproducing Kim
et al.~2022); point-AUC of random = 0.496 ≈ chance; best-affiliation-F1
of random = 0.681 (independently reproducing SimAD's ≈0.7); VUS-ROC with
buffer→1 equals point-AUC exactly, and its random-null rises
monotonically with the buffer window (a characterized property, not an
artifact); PA\%K deflates random (0.90→0.28→0.18); an oracle detector
scores VUS-ROC 0.994 ≫ random. This certifies that any gameability we
report is a metric property, not a harness bug.

\hypertarget{data}{%
\subsection{Data}\label{data}}

\textbf{Primary: UCR Anomaly Archive} (Wu \& Keogh 2021), 250 univariate
series (verified: length median 30k, range 6.7k--900k; anomaly length
median 101; prevalence median 0.34\%), one anomaly per series by
construction. Series longer than 40,000 points are cropped to a 40k
window centered on the anomaly (a timing-driven cap retaining ample
normal context); this bounds VUS cost without affecting the gameability
question. \textbf{Secondary (now reported in §5.5):} SMD (Su et
al.~2019; 28 series), SMAP/MSL (Hundman et al.~2018; 40/22), NAB (Lavin
\& Ahmad 2015; 52), and PSM (Abdulaal et al.~2021; a single train/test
split → 1 series, used only qualitatively as it cannot support a rate),
via TSB-AD mirrors. Median anomaly prevalence on these datasets ranges
2.1--10.0\%, one to two orders of magnitude above UCR's 0.34\%, letting
us test whether the findings survive outside UCR's extreme
low-prevalence, single-anomaly regime.

\hypertarget{configuration-frozen}{%
\subsection{Configuration (frozen)}\label{configuration-frozen}}

Length cap 40k; R=10 replicates; VUS threshold resolution 250;
gameability fraction 0.90; near-miss shifts \{0.1,\ldots,4\}×median
anomaly length; seed 20260706. All locked in a released,
version-controlled configuration file, released with the harness (see
Data availability).

\hypertarget{results}{%
\section{Results}\label{results}}

Results are over all 250 UCR Anomaly Archive series (ids 001--250,
complete; one byte-identical duplicate file deduplicated). All
gameability figures use the PURE no-skill pool (no anomaly-location
information); the oracle-inclusive rate (adding the
single-spike-inside-anomaly detector) is reported alongside as an upper
reference.

\begin{figure}
\centering
\includegraphics[width=0.9\textwidth,height=\textheight]{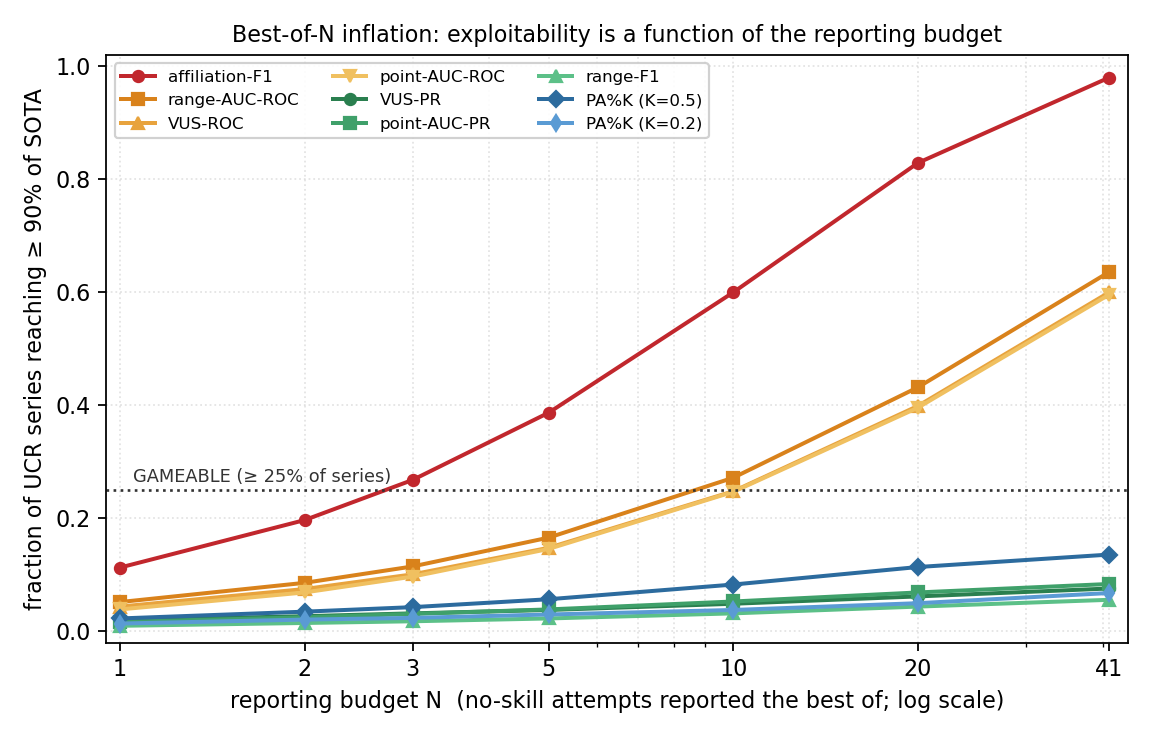}
\caption{Best-of-N inflation (the paper's central result): the fraction
of UCR series on which a pure no-skill adversary reaches at least 90\%
of the SOTA reference, as a function of N, the number of random attempts
reported the best of. At N=1 no metric is gameable; affiliation-F1 and
the ROC-based metrics inflate steeply with N while the PR-based metrics
and PA\%K stay floored at the anomaly prevalence. Dotted line = the
GAMEABLE threshold (25\%).}
\end{figure}

\hypertarget{the-best-of-n-inflation-curve}{%
\subsection{The best-of-N inflation
curve}\label{the-best-of-n-inflation-curve}}

Figure 1 is the paper's central result: for each replacement metric, the
fraction of UCR series on which a pure no-skill adversary reaches ≥90\%
of the SOTA reference, as a function of the reporting budget N (Sec
3.3). Table 2 reads three operating points off that curve.

\textbf{At N=1 no replacement metric is gameable.} A single random
no-skill attempt reaches 90\% of an excellent real-detector SOTA on at
most 11\% of series, and the three-way split is already visible: the
PR-based metrics and PA\%K are ROBUST (0.01--0.02 of series), the ROC
family is near-ROBUST (0.04--0.05), and affiliation-F1 is already
FRAGILE (0.11 for a random draw, and up to 0.19 for the single worst
trivial detector). Against one honest run of one no-skill detector,
then, not one of the eleven replacement metrics is \emph{gameable} (none
reaches the 25\% GAMEABLE line), which is what a replacement for
gameable point-adjustment should deliver, with the single caveat that
affiliation is already mildly exploitable, from the intrinsic leniency
we return to below.

\textbf{The metrics diverge sharply as N grows} (Figure 1, Table 2), and
the axis is ROC/affiliation vs PR. affiliation-F1 crosses the GAMEABLE
line (25\%) by \textbf{N=3} and reaches \textbf{0.98} at the full pool
(N=41): a best-of-three random detector already looks state-of-the-art
on a quarter of series. The ROC family (VUS-, point-, range-AUC-ROC)
crosses GAMEABLE by \textbf{N=9--11} and reaches \textbf{0.60--0.64} at
N=41. The PR family (VUS-PR, point-AUC-PR, range-F1) and PA\%K
\textbf{never reach GAMEABLE at any N ≤ 41 on UCR}: they plateau at
\textbf{0.06--0.14} (the one benchmark where a PR metric does tip is NAB
at large N, §5.5). Best-of-N reporting inflates the ROC and affiliation
metrics steeply and the PR metrics not at all. At the seed-shopping
worst case (N=41, the adversarial best over all attempts, the statistic
prior proposer-run checks implicitly used), the split is decisive: a
paired McNemar test finds VUS-ROC inflated on 131 series where its
sibling VUS-PR is not, and the reverse on \textbf{zero} (p ≈ 7×10⁻⁴⁰);
point-AUC is one-directional too (129 vs 1). All four ROC/affiliation
``GAMEABLE-at-N=41'' calls survive Holm correction across the twelve
metrics; no PR/PA\%K metric does at any N.

\emph{Table 2: inflation rate at three reporting budgets (pure no-skill,
Jeffreys 95\% CIs; n=250). N* = the smallest N at which the metric
crosses GAMEABLE (25\%), ``---'' = never within N≤41. med = the median
value a single random draw achieves (the leniency baseline). Class =
frozen-ladder tier at the N=41 worst case (the full pure pool).
range-AUC-PR is omitted here (its GAMEABLE label is a floored-reference
artifact, §5.4); point-wise F1 tracks the PR family (0.10 at N=41,
FRAGILE). PA-F1 (the pre-fix control) is GAMEABLE from N≈1.}

\begin{longtable}[]{@{}lcccccl@{}}
\toprule
Metric & rate @N=1 & rate @N=10 & rate @N=41 & N* & med 1-draw &
Class@N=41 \\
\midrule
\endhead
affiliation-F1 & 0.11 & 0.60 & \textbf{0.98} & \textbf{3} & 0.68 &
GAMEABLE \\
range-AUC-ROC & 0.05 & 0.27 & 0.64 & 9 & 0.62 & GAMEABLE \\
point-AUC-ROC & 0.04 & 0.25 & 0.60 & 11 & 0.50 & GAMEABLE \\
VUS-ROC & 0.04 & 0.25 & 0.60 & 11 & 0.60 & GAMEABLE \\
PA\%K (K=0.5) & 0.02 & 0.08 & 0.14 & --- & 0.02 & FRAGILE \\
point-AUC-PR & 0.02 & 0.05 & 0.08 & --- & 0.006 & FRAGILE \\
VUS-PR & 0.02 & 0.05 & 0.08 & --- & 0.009 & FRAGILE \\
PA\%K (K=0.2) & 0.01 & 0.04 & 0.07 & --- & 0.05 & FRAGILE \\
range-based F1 & 0.01 & 0.03 & 0.06 & --- & 0.009 & FRAGILE \\
\bottomrule
\end{longtable}

\textbf{Why best-of-N inflates ROC and affiliation but not PR
(mechanism).} The three families sit at very different single-draw
baselines (Table 2, ``med 1-draw''), and best-of-N is a monotone order
statistic that lifts a metric toward SOTA only when its single-draw
distribution has both a high mean and real variance. (i) \emph{AUC-ROC
of a random score against one rare, contiguous anomaly is a
high-variance variable centred near 0.5}: the effective positive sample
size is one segment, so a random ranking's ROC-AUC scatters widely
(median single draw 0.50--0.62 here); reporting the best of N draws
walks that scatter up to 0.9. (ii) \emph{AUC-PR is floored at the
anomaly prevalence} (median 0.34\% on UCR): a random score's precision
cannot exceed the base rate, so its PR-AUC sits at ≈0.01 with little
variance (median single draw 0.006--0.009) and best-of-N has nothing to
climb; this is the classical ROC-vs-PR behaviour under extreme class
imbalance (Davis \& Goadrich 2006; Saito \& Rehmsmeier 2015). (iii)
\emph{affiliation-F1 is extremely lenient}: a single random score
already achieves a median 0.68 (chance 0.75, Sec 5.4), because
affiliation rewards mere temporal proximity and forgives misplacement of
4× the anomaly length (Sec 5.2); starting two-thirds of the way to a
perfect score, it is already fragile at N=1 and best-of-three suffices
to clear 90\% on a quarter of series. The PR family's prevalence floor
is thus a near-total resistance to best-of-N reporting (NAB the one
exception, §5.5) that the ROC and affiliation families lack.

Two consequences follow. First, a raw no-skill number reported without
its N is uninterpretable: the same random detector is not gameable at
N=1 and GAMEABLE at N≥3--11 under ROC/affiliation. Second,
affiliation-F1 under best-of-N is worse than the PA-F1 control it was
meant to improve upon (0.98 vs PA-F1's ≈0.76 at the same N): the field
replaced a metric it deemed catastrophically permissive with one that,
under seed-shopping, is more so.

\begin{figure}
\centering
\includegraphics[width=0.9\textwidth,height=\textheight]{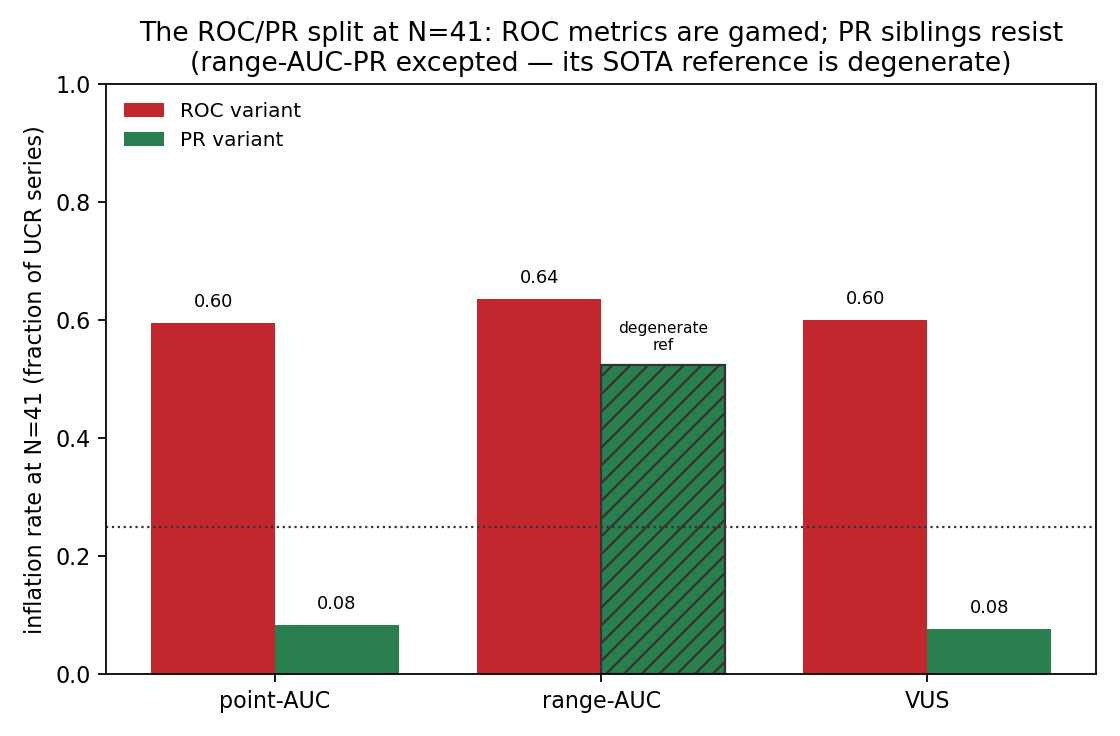}
\caption{The ROC/PR split at the N=41 (full pure pool) best-of-N worst
case: each ROC metric's inflation rate beside its PR sibling
(range-AUC-PR excepted, its SOTA reference is degenerate, Sec. 5.4).}
\end{figure}

\hypertarget{forgiveness-frontier}{%
\subsection{Forgiveness frontier}\label{forgiveness-frontier}}

Figure 3 reports the near-miss frontier: the largest ground-truth shift
(in median anomaly lengths) at which a zero-localization detector still
reaches ≥90\% of SOTA (bootstrap 95\% CIs on each point). The metrics
span an order of magnitude in spatial tolerance. \textbf{affiliation-F1
forgives a detector misplaced by 4× the median anomaly length} (its
median near-miss ratio stays ≥0.98 out to the largest shift tested), a
quantitative statement of why it is so gameable. The ROC-family
(point/range/VUS-ROC) forgives 2×; every remaining metric (the
PA-family, the PR-based metrics, point-wise and range-F1) forgives only
0.5×, punishing anomaly-scale misplacement. Spatial forgiveness tracks
gameability rank closely, identifying it as the mechanism: the metrics
that tolerate misplacement are exactly the ones no-skill detectors
exploit, and affiliation's extreme tolerance (its median single random
score is already 0.68) is what lets a best-of-three no-skill run already
clear the GAMEABLE line and best-of-N push it on to SOTA.

\begin{figure}
\centering
\includegraphics[width=0.9\textwidth,height=\textheight]{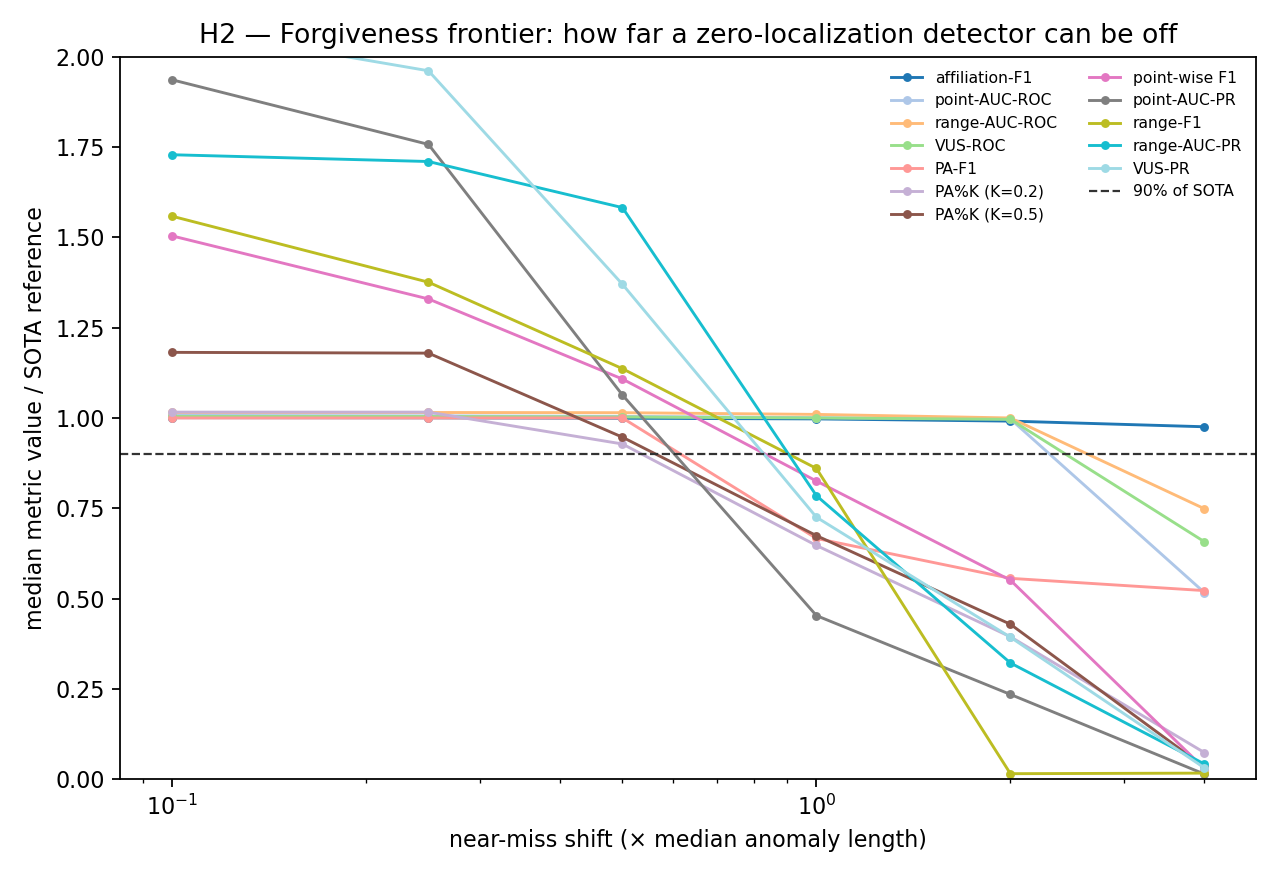}
\caption{Forgiveness frontier: median near-miss ratio-to-SOTA vs
ground-truth shift (in median anomaly lengths). affiliation-F1 stays at
SOTA out to 4×.}
\end{figure}

\hypertarget{inter-metric-rank-agreement}{%
\subsection{Inter-metric rank
agreement}\label{inter-metric-rank-agreement}}

Over the six-detector pool, the mean per-series pairwise Kendall τ
between metrics is \textbf{0.684} (median 0.71): \textbf{metrics largely
agree} on how they rank detectors. The fraction of series with
τ\textless0.6 is \textbf{0.240} (Figure 4), which does \textbf{not}
exceed the 25\% threshold (one-sided exact binomial p = 0.67; Jeffreys
95\% CI {[}0.19, 0.30{]}); restricted to the replacement metrics it is
0.180, and among the recommended survivors 0.104 (mean τ 0.78).
\textbf{Rank agreement is therefore not established:} that CI straddles
the 25\% threshold, so we fail to establish aggregate reordering rather
than demonstrate its absence (an equivalence claim would need a TOST the
CI cannot support). The one caveat is at the very top: the single
highest-ranked detector differs across metrics on a majority of series
(top-1 flip 0.63 among survivors, 0.84 among replacements), so the
\emph{identity of the best detector} is metric-sensitive even where the
overall ordering concurs. With only six detectors this top-1 statistic
is coarse and we report it as indicative. The disagreement that does
occur concentrates on the ROC/PR axis.

\begin{figure}
\centering
\includegraphics[width=0.9\textwidth,height=\textheight]{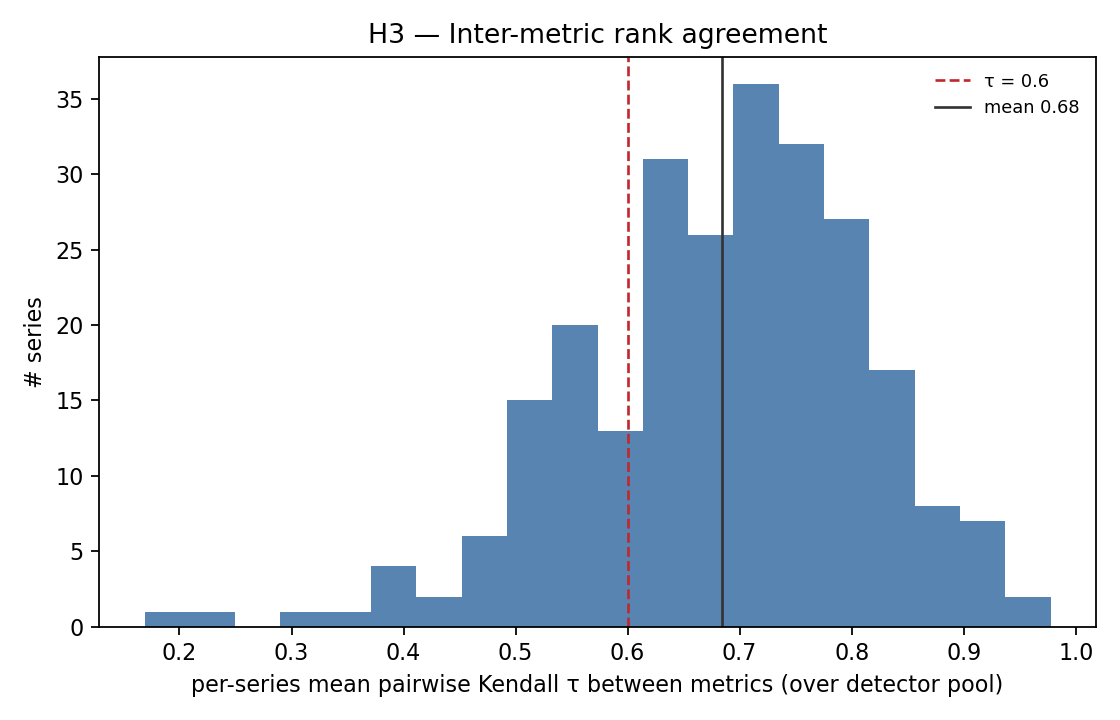}
\caption{Inter-metric rank agreement: distribution of per-series mean
pairwise Kendall τ.}
\end{figure}

\hypertarget{robustness-the-weak-reference-objection-resolved-in-both-directions}{%
\subsection{Robustness: the weak-reference objection, resolved in both
directions}\label{robustness-the-weak-reference-objection-resolved-in-both-directions}}

The central threat to a SOTA-relative audit is a \emph{weak} SOTA
reference: if the real detectors barely beat chance, best-of-N reaching
90\% of them is trivial. We test this with the \emph{gamed-series SOTA}
(gSOTA, the median real-detector SOTA on precisely the series a metric
reaches under best-of-N), applied evenhandedly to every metric so the
diagnostic cannot be selectively invoked.

\emph{The best-of-N inflation is genuine, not a weak-reference
artifact.} On the very series where best-of-N reaches SOTA under
affiliation and the ROC metrics, the real detectors themselves score
\textbf{0.97--1.00} (gSOTA); a strong SOTA is below 0.6 on only 0--3\%
of those series. A stronger reference pool therefore cannot rescue those
metrics: SOTA is already at the ceiling exactly where best-of-N matches
it. Every metric also separates real detectors from chance by a wide
margin (median SOTA − median chance = 0.25 for affiliation, 0.37--0.49
for the ROC metrics; Table 3).

\emph{The one exception, range-AUC-PR, is a degenerate reference, not a
genuine failure.} range-AUC-PR is PR-based yet GAMEABLE (0.516), which
would break the ROC/PR dichotomy. But its gSOTA is \textbf{0.19}, and on
\textbf{100\%} of the series it ``games'', the real detectors score
below 0.6: range-AUC-PR simply assigns near-zero to every detector, real
or fake (median no-skill 0.51 ≈ median SOTA 0.51), so the ratio is
meaningless. Its GAMEABLE label is a floored-reference artifact,
qualitatively different from the genuine ROC/affiliation gaming, and we
exclude it from the ``PR is safe'' recommendation. The FRAGILE PR-family
(VUS-PR, point-AUC-PR, range-F1) shows a milder version of the same
effect: their small residual gaming occurs on near-floor-SOTA series
(gSOTA 0.04--0.17), so their true robustness is likely \emph{better}
than the 0.06--0.08 reported.

\emph{Fixed-threshold sensitivity confirms it.} Removing the oracle
best-F1 threshold and predicting exactly the anomaly budget (a fixed
quantile), the same ordering holds: affiliation-F1 is still gamed on
\textbf{66\%} of series, while point-wise F1 and range-F1 drop to
\textbf{0.09 and 0.08}; the PR/point family is \emph{even more} robust
without the oracle threshold, and affiliation's gameability is
\emph{not} an artifact of it.

\emph{Cutoff-invariance at fixed N, but N is the real axis.} The two
scalar knobs (the 90\%-of-SOTA fraction and the 25\% GAMEABLE cutoff) do
not drive the N=41 classification: recomputing at fractions
0.85/0.90/0.95 and cutoffs 20/25/30\% leaves every tier unchanged (the
ROC/affiliation rates sit at 0.28 or above, the PR/PA\%K rates at 0.14
or below, so nothing lies near a boundary), and the VUS-ROC-vs-VUS-PR
McNemar split is one-directional at every fraction (discordant b =
162/131/54, c = 0/0/1, p \textless{} 10⁻¹⁴). But the decisive robustness
result is not cutoff-invariance at a fixed N; it is the full N-curve
(Sec 5.1), whose N=1 baseline shows no metric is gameable there and
whose ROC-over-PR ordering is what transfers across benchmarks. The
audit's credibility rests on that single-run baseline, on independence
(we propose none of the metrics), and on full reproducibility (every
number regenerates from the released harness by one command; the
configuration and the sensitivity sweep are released, see Data
availability).

\emph{The adversary is uninformative structure.} The most effective
adversary is the \textbf{random-walk} score for the ROC and FRAGILE
metrics, and plain \textbf{i.i.d.\ noise} for affiliation and PA-F1
(Table 3). Both are anomaly-blind; a random walk merely mimics the
smooth texture of a real detector's output. These metrics reward
\emph{score smoothness or spread that overlaps buffered anomaly
regions}, which no-skill signals possess for free.

\emph{Table 3: SOTA vs chance separation and no-skill exploitation
(medians over 250 series). Every metric separates real detectors from
chance (positive separation); the GAMEABLE metrics differ in that the
best no-skill generator also reaches SOTA.}

\begin{longtable}[]{@{}lccccl@{}}
\toprule
Metric & median SOTA & median chance & separation & median no-skill &
top adversary \\
\midrule
\endhead
affiliation-F1 & 1.00 & 0.75 & 0.25 & 0.98 & i.i.d.\ noise \\
PA-F1 & 1.00 & 0.65 & 0.35 & 0.98 & i.i.d.\ noise \\
range-AUC-ROC & 0.98 & 0.62 & 0.37 & 0.92 & random-walk \\
point-AUC-ROC & 0.99 & 0.50 & 0.49 & 0.90 & random-walk \\
VUS-ROC & 1.00 & 0.60 & 0.39 & 0.92 & random-walk \\
range-AUC-PR & 0.51 & 0.01 & 0.50 & 0.51 & constant \\
PA\%K (K=0.5) & 0.85 & 0.02 & 0.82 & 0.10 & random-walk \\
point-wise F1 & 0.61 & 0.02 & 0.59 & 0.08 & random-walk \\
point-AUC-PR & 0.51 & 0.01 & 0.50 & 0.03 & random-walk \\
VUS-PR & 0.47 & 0.01 & 0.46 & 0.03 & random-walk \\
range-based F1 & 0.54 & 0.01 & 0.53 & 0.03 & random-walk \\
PA\%K (K=0.2) & 0.98 & 0.06 & 0.93 & 0.16 & random-walk \\
\bottomrule
\end{longtable}

\hypertarget{cross-benchmark-replication-five-further-benchmarks}{%
\subsection{Cross-benchmark replication (five further
benchmarks)}\label{cross-benchmark-replication-five-further-benchmarks}}

To test whether the UCR findings are specific to that archive's extreme
low-prevalence, single-anomaly structure, we re-ran the identical
pipeline on five further benchmarks accessed through TSB-AD mirrors: SMD
(28 series), SMAP (40), MSL (22), NAB (52), and PSM (1 series, a single
train/test split, used only qualitatively as it cannot support a rate).
Median anomaly prevalence rises from UCR's 0.34\% to 2.1--10.0\%. We
report both the N=1 single-run rate and, in Table 4, the best-of-N worst
case (the deterministic best over each series' full 41-attempt pure
pool). Computed from the same pools as §5.1, Table 4's UCR column
reproduces the §5.1 N=41 column exactly, a built-in positive control
that the multi-benchmark harness is correct.

\emph{Table 4: Cross-benchmark inflation rate at the best-of-N worst
case (best of the 41 pure no-skill attempts reaches ≥90\% of
best-in-pool SOTA), per metric × benchmark, with the frozen-ladder class
(G = GAMEABLE ≥25\%, F = FRAGILE 5--25\%, R = ROBUST ≤5\%). At N=1 no
metric is gameable on any benchmark (≤0.12). Benchmarks are ordered by
median anomaly prevalence (in parentheses), which is deliberately
non-monotonic in the rates. Verdict: REPLICATES = same class on all five
rate-bearing benchmarks; MOSTLY = same class on all but one; VARIES =
spans two or more classes. PA-F1 (control) and PSM (1 series) are
omitted from the table. †range-AUC-PR is the degenerate-reference
exception (§5.4).}

\begin{longtable}[]{@{}lcccccl@{}}
\toprule
Metric & UCR (0.34\%) & SMAP (2.1\%) & SMD (3.3\%) & MSL (6.6\%) & NAB
(10.0\%) & Verdict \\
\midrule
\endhead
affiliation-F1 & 0.98 G & 0.85 G & 0.07 F & 0.77 G & 0.60 G & MOSTLY \\
point-AUC-ROC & 0.60 G & 0.42 G & 0.18 F & 0.41 G & 0.60 G & MOSTLY \\
range-AUC-ROC & 0.64 G & 0.45 G & 0.00 R & 0.64 G & 0.71 G & MOSTLY \\
VUS-ROC & 0.60 G & 0.42 G & 0.00 R & 0.41 G & 0.62 G & MOSTLY \\
range-AUC-PR† & 0.52 G & 0.15 F & 0.32 G & 0.27 G & 0.63 G & MOSTLY \\
PA\%K (K=0.2) & 0.07 F & 0.10 F & 0.07 F & 0.14 F & 0.29 G & MOSTLY \\
PA\%K (K=0.5) & 0.14 F & 0.10 F & 0.14 F & 0.14 F & 0.25 G & MOSTLY \\
point-wise F1 & 0.10 F & 0.05 R & 0.14 F & 0.09 F & 0.21 F & MOSTLY \\
point-AUC-PR & 0.08 F & 0.05 R & 0.14 F & 0.05 R & 0.13 F & VARIES \\
range-based F1 & 0.06 F & 0.15 F & 0.00 R & 0.23 F & 0.35 G & VARIES \\
VUS-PR & 0.08 F & 0.05 R & 0.04 R & 0.09 F & 0.25 G & VARIES \\
\bottomrule
\end{longtable}

The single-run result and the best-of-N direction both replicate; the
magnitude is benchmark-dependent.

\emph{(i) At N=1 no metric is gameable, on any benchmark.} On every one
of the five further benchmarks, as on UCR, the N=1 inflation rate is
≤0.11 for \emph{every} metric (the PR family and PA\%K ≤0.05;
affiliation and the ROC family up to 0.11 on the low-prevalence
benchmarks), ROBUST-or-FRAGILE everywhere, nothing GAMEABLE. That one
honest run games no replacement metric is not a UCR artifact: it holds
on all six benchmarks.

\emph{(ii) The best-of-N ROC-over-PR inflation replicates on the
higher-prevalence benchmarks.} Table 4 gives the best-of-N worst case.
On SMAP and MSL the pattern is clean (affiliation and every ROC metric
inflate to GAMEABLE, 0.41--0.85, while the PR family stays
FRAGILE-or-ROBUST), and the VUS-ROC-vs-VUS-PR McNemar split never
reverses on any benchmark (discordant \emph{b} = 131/15/7/19 on
UCR/SMAP/MSL/NAB, \emph{c} = 0/0/0/0; p from 7×10⁻⁴⁰ to 2×10⁻²), null
only on SMD (b=0). The direction (ROC/affiliation inflatable under
best-of-N, PR resistant) is therefore not a UCR artifact. range-AUC-PR
is GAMEABLE on four benchmarks via its degenerate reference (§5.4), a
floored-reference artifact rather than genuine gaming.

\emph{(iii) Two honest exceptions bound the claim.} (a) \textbf{SMD}
shows almost no inflation even at N=41 (affiliation 0.07, the ROC family
0.00--0.18): it is the one benchmark on which the in-house detector pool
itself scores highest, lifting SOTA beyond even best-of-N reach, so the
audit correctly reports little inflation exactly where the real
detectors are genuinely strong. (b) \textbf{NAB} is the hardest
benchmark, and at N=41 even the PR family tips GAMEABLE there (VUS-PR
0.25, PA\%K 0.25--0.29, range-F1 0.35). So under best-of-N reporting
\textbf{no single metric is safe on all six benchmarks}: the PR family's
immunity is near-total but not absolute, breaking on NAB. The
transferable finding is the ordering (under best-of-N, affiliation and
the ROC family inflate first and furthest, PR and PA\%K last and least),
while the budget N at which a given metric becomes exploitable shifts
with the benchmark.

\hypertarget{discussion}{%
\section{Discussion}\label{discussion}}

\textbf{Did the fix work? For a single run, yes; under best-of-N
reporting, not for ROC or affiliation.} Against one honest run of one
no-skill detector, no post-point-adjustment replacement metric is
gameable: at N=1 none reaches the GAMEABLE line on any benchmark (the PR
family and PA\%K are inflated on ≤2\% of UCR series, affiliation and the
ROC family on ≤11\%), and this holds on all six benchmarks (Sec 5.1,
5.5). The fix delivered what a replacement for gameable point-adjustment
should. The failure is \textbf{conditional on reporting practice}. An
adversary who runs a random detector repeatedly and reports only the
best seed (the \emph{best-of-N seed-shopping} endemic to empirical ML),
inflates affiliation-F1 to state-of-the-art on 98\% of series by N=41
(past GAMEABLE already by N=3) and every ROC-based metric past GAMEABLE
by N≈9--11, while the PR-based metrics and PA\%K stay flat at every N.
The mechanism is the classical order-statistic behaviour of AUC under
extreme class imbalance (Sec 5.1; Davis \& Goadrich 2006; Saito \&
Rehmsmeier 2015): a random ROC-AUC has the variance for best-of-N to
climb toward SOTA, a random PR-AUC is floored at the anomaly prevalence
and cannot, and affiliation starts two-thirds of the way to a perfect
score. Crucially the danger is not evenly distributed within a family:
\textbf{VUS-ROC and VUS-PR are proposed together, ship in the same
package, and are reported interchangeably, yet under best-of-N one
inflates on 131 series and the other on none of those same series.} The
one apparent exception, range-AUC-PR, is a floored-reference artifact
(§5.4), not genuine PR-metric exploitability.

\textbf{A metric-selection recommendation.} From the ladder, the
gamed-SOTA diagnostic, the frontier, and the cross-benchmark table we
distill a decision protocol (released with the harness): \emph{disclose
the number of random seeds N behind any reported score, and treat a
no-skill or best-of-N-reported ROC-AUC or affiliation number as
uninterpretable without its N; prefer a PR-based F1 or AUC-PR metric
(VUS-PR foremost, already the TSB-AD headline) or PA\%K, which resist
best-of-N inflation on nearly every benchmark (NAB at large N excepted);
report any ROC-AUC variant only alongside a PR counterpart; treat
affiliation-F1 as inflatable unless paired with its forgiveness-frontier
value; never read range-AUC-PR without checking that the real detectors
score meaningfully above the prevalence floor; and, because the budget N
at which a metric becomes exploitable is benchmark-dependent (§5.5),
verify on your own data.} This \textbf{supports the field's recent shift
toward the PR family as the least-inflatable class}, with the honest
caveat that even VUS-PR, resistant on UCR/SMD/SMAP/MSL (≤0.09), tips
GAMEABLE on NAB at large N. The transferable guarantee is the
\emph{ordering} under best-of-N; a single metric's absolute safety does
not survive the change of benchmark. The fix closed single-run
gameability but not best-of-N inflation of the ROC and affiliation
metrics.

\textbf{Why independence matters here.} The single most reassuring prior
fact about these metrics, that they are robust, traces almost entirely
to their proposers. Our audit is adversarial (we \emph{try} to game each
metric), SOTA-relative (we ask whether no-skill reaches real detectors,
not merely whether it beats chance), independent (we propose none of the
metrics), and explicit about the reporting budget: at N=1 no metric is
gameable, the ROC-over-PR inflation appears only under best-of-N, and it
replicates across benchmarks (§5.1, §5.5). Under those conditions the
reassuring picture holds for the PR metrics and breaks, under best-of-N,
for the ROC and affiliation metrics, a distinction the proposer-run
checks, which report a single adversarial-best number at one fixed N,
were not designed to draw.

\textbf{Reconciliation with proposer claims.} Our results do not
contradict the metric papers: affiliation \emph{is} designed to reward
temporal proximity, and VUS \emph{is} threshold- and buffer-robust. What
we add is that, measured adversarially and relative to SOTA on real
data, proximity-reward becomes no-skill-exploitability for the ROC
variants and for affiliation, while the PR variants' precision term
suppresses it. The forgiveness frontier makes this a quantity, not a
verdict: it is the price, in anomaly-lengths of tolerated misplacement,
that each metric pays for its robustness or leniency.

\hypertarget{limitations}{%
\section{Limitations}\label{limitations}}

\begin{enumerate}
\def\labelenumi{\arabic{enumi}.}
\tightlist
\item
  \textbf{SOTA reference is an in-house six-detector pool} (Matrix
  Profile, IF, LOF, AR, PCA-reconstruction, kNN-distance). The
  gamed-SOTA diagnostic (§5.4) shows this does \emph{not} inflate the
  headline ROC/affiliation result (real detectors score 0.97--1.0 on the
  gamed series) but \emph{does} contaminate range-AUC-PR and, mildly,
  the FRAGILE PR-family (near-floor SOTA on their few gamed series).
  Wiring published-leaderboard detector outputs as an alternative
  reference is the natural next strengthening.
\item
  \textbf{The no-skill statistic is best-of-N}, an order statistic in
  the reporting budget N. Rather than fix a single N and risk mistaking
  an order-statistic artifact for a metric property, we report the full
  N-curve (§5.1): the N-dependence is the finding, not a confound, and
  the N=1 point isolates genuine single-run behaviour. Best-F1
  thresholding grants each detector its oracle threshold, applied
  symmetrically to no-skill and SOTA; the fixed-threshold sensitivity
  (§5.4) shows the ROC/PR ordering is threshold-independent.
\item
  \textbf{Benchmark scope.} We now report five secondary benchmarks
  (§5.5): the ROC/PR \emph{direction} replicates, but \emph{absolute}
  rates are benchmark-dependent, and one benchmark (SMD) shows little
  gaming because its real detectors are strong there, so a metric's tier
  must be verified per dataset, not assumed. The secondary series counts
  are modest (22--52) and PSM is a single series (qualitative only);
  industrial-control coverage (SWaT/WADI) and a published-leaderboard
  SOTA reference remain the next strengthening.
\item
  \textbf{Rank agreement uses a six-detector pool} → the rank-agreement
  τ is coarse, and the top-1-flip statistic especially so; we report it
  as \emph{not established} rather than over-reading it.
\item
  \textbf{The near-miss battery invites a ``tolerance is by design''
  defense} from metric authors; the frontier in anomaly-length units is
  our pre-emptive, quantitative answer, not a refutation of the metrics'
  intent.
\item
  Prior work reaches adjacent conclusions: notably Yang et al.~(2025),
  which predates this study by about eight months but is
  proposer-conflicted (SimAD authorship) and not SOTA-relative. Our
  contribution is the outside-proposer, SOTA-relative adversarial frame
  plus the released harness, not primacy on the bare fact that no-skill
  inflates these metrics.
\end{enumerate}

\hypertarget{conclusion}{%
\section{Conclusion}\label{conclusion}}

\textbf{The point-adjustment fix works for a single run but not for
best-of-N reporting.} We provide the first independent, adversarial,
SOTA-relative stress-test of the post-point-adjustment TSAD metric suite
on real benchmarks, and identify the reporting budget N (how many random
attempts an adversary reports the best of) as the axis that governs
metric robustness. At N=1, one honest run of one no-skill detector, no
replacement metric is gameable on any of the six benchmarks: the fix
delivered single-run robustness (with affiliation already mildly
fragile, from its intrinsic leniency). But under best-of-N reporting,
the seed-shopping common in empirical ML, affiliation-F1 and every
ROC-based metric inflate to SOTA (affiliation crossing GAMEABLE by N=3
and reaching 98\% by N=41, the ROC family by N≈9--11), while the
PR-based metrics and PA\%K stay flat, floored at the anomaly prevalence.
The split is decisive and one-directional (VUS-ROC inflated on 131
series where VUS-PR is not, never the reverse), replicates across
benchmarks, and follows from the order-statistic behaviour of AUC-ROC
versus AUC-PR under extreme class imbalance (affiliation inflates by a
second route, its extreme single-run leniency). The practical verdict is
a rule the field can act on today: \textbf{disclose the number of seeds
N behind any reported score, prefer PR-based metrics (which resist
best-of-N inflation on nearly every benchmark) or PA\%K, and treat any
best-of-N-reported ROC-AUC or affiliation number as uninterpretable
without its N.} The field's move toward PR-based scoring is supported as
the least-inflatable class; its still-common interchangeable reporting
of VUS-ROC and affiliation-F1 is not.

\hypertarget{data-and-code-availability}{%
\section{Data and code availability}\label{data-and-code-availability}}

UCR Anomaly Archive is public (Wu \& Keogh 2021). Our stress-test
harness, all metric vendors at frozen commits (affiliation @8d84498,
TSB-UAD @313f0fd, prts 1.0.0.3), the frozen config, the raw per-series
results for all six benchmarks, and one-command reproduction scripts are
provided with this submission as supplementary material for review, so
every reported number regenerates from the shipped results without
re-running the grid; the archive will receive a DOI on acceptance. The
released configuration fixes every outcome-defining choice (the ladder
thresholds, metric-inclusion rules, no-skill generators, thresholding
protocol, and statistical tests); any deviation from it is disclosed in
the text where it occurs (for example, the detector-pool expansion,
§5.4). The audit's robustness is established directly by the full
N-curve (§5.1, whose N=1 baseline isolates single-run behaviour) and the
cross-benchmark replication (§5.5).

\hypertarget{declaration-of-generative-ai-use}{%
\section{Declaration of Generative AI
use}\label{declaration-of-generative-ai-use}}

During the preparation of this work the author used Claude (Anthropic)
as an assistive tool: to help draft and edit prose, to help implement
the analysis and experiment code, and to help run the experiment
pipeline. The study's design, structure, and framing; the topic and
hypotheses; every analysis decision (set out in the released,
version-controlled configuration); the review and revision of all
content; and the verification of every result are the author's, who
takes full responsibility for the paper. Every reported number is
regenerable from the released logs and code by a single command; the
author independently re-derived the headline gameability rates from the
raw result files.

\hypertarget{declarations}{%
\section{Declarations}\label{declarations}}

\textbf{Funding.} No funding was received for this work.

\textbf{Competing interests.} The author declares no competing
interests.

\textbf{Data availability.} See ``Data and code availability'' above;
the UCR Anomaly Archive is public, and the harness and the frozen
configuration are released (provided with this submission).

\hypertarget{references}{%
\section*{References}\label{references}}

Abdulaal, A., Liu, Z., Lancewicki, T. (2021). A Practical Approach to
Asynchronous Multivariate Time Series Anomaly Detection and
Localization. \emph{KDD}. doi:10.1145/3447548.3467174.

Bhattacharya, D., Mukherjee, S., Kamanchi, C., Ekambaram, V., Jati, A.,
Dayama, P. (2024). Towards Unbiased Evaluation of Time-series Anomaly
Detector. arXiv:2409.13053.

Breunig, M. M., Kriegel, H.-P., Ng, R. T., Sander, J. (2000). LOF:
Identifying Density-Based Local Outliers. \emph{ACM SIGMOD}.
doi:10.1145/342009.335388.

Davis, J., Goadrich, M. (2006). The Relationship Between
Precision-Recall and ROC Curves. \emph{ICML}, 233--240.
doi:10.1145/1143844.1143874.

Garg, A., Zhang, W., Samaran, J., Savitha, R., Foo, C.-S. (2022). An
Evaluation of Anomaly Detection and Diagnosis in Multivariate Time
Series. \emph{IEEE TNNLS} 33(6):2508--2517.
doi:10.1109/TNNLS.2021.3105827.

Ghorbani, R., Reinders, M. J. T., Tax, D. M. J. (2024). PATE:
Proximity-Aware Time Series Anomaly Evaluation. \emph{KDD}.
doi:10.1145/3637528.3671971.

Huet, A., Navarro, J. M., Rossi, D. (2022). Local Evaluation of Time
Series Anomaly Detection Algorithms. \emph{KDD}, 635--645.
doi:10.1145/3534678.3539339.

Hundman, K., Constantinou, V., Laporte, C., Colwell, I., Soderstrom, T.
(2018). Detecting Spacecraft Anomalies Using LSTMs and Nonparametric
Dynamic Thresholding. \emph{KDD}. doi:10.1145/3219819.3219845.

Hwang, W.-S., Yun, J.-H., Kim, J., Min, B. G. (2022). Do you know
existing accuracy metrics overrate time-series anomaly detections?
\emph{ACM SAC}, 403--412. doi:10.1145/3477314.3507024.

Kim, S., Choi, K., Choi, H.-S., Lee, B., Yoon, S. (2022). Towards a
Rigorous Evaluation of Time-series Anomaly Detection. \emph{AAAI}
36(7):7194--7201. doi:10.1609/aaai.v36i7.20680.

Lavin, A., Ahmad, S. (2015). Evaluating Real-Time Anomaly Detection
Algorithms --- The Numenta Anomaly Benchmark. \emph{IEEE ICMLA}, 38--44.
doi:10.1109/ICMLA.2015.141.

Liu, F. T., Ting, K. M., Zhou, Z.-H. (2008). Isolation Forest.
\emph{IEEE ICDM}, 413--422. doi:10.1109/ICDM.2008.17.

Liu, Q., Paparrizos, J. (2024). The Elephant in the Room: Towards A
Reliable Time-Series Anomaly Detection Benchmark. \emph{NeurIPS Datasets
\& Benchmarks}. doi:10.52202/079017-3437.

Paparrizos, J., Boniol, P., Palpanas, T., Tsay, R. S., Elmore, A.,
Franklin, M. J. (2022). Volume Under the Surface: A New Accuracy
Evaluation Measure for Time-Series Anomaly Detection. \emph{PVLDB}
15(11):2774--2787. doi:10.14778/3551793.3551830.

Paparrizos, J., Kang, Y., Boniol, P., Tsay, R. S., Palpanas, T.,
Franklin, M. J. (2022). TSB-UAD: An End-to-End Benchmark Suite for
Univariate Time-Series Anomaly Detection. \emph{PVLDB} 15(8):1697--1711.
doi:10.14778/3529337.3529354.

Rewicki, F., Denzler, J., Niebling, J. (2023). Is It Worth It? Comparing
Six Deep and Classical Methods for Unsupervised Anomaly Detection in
Time Series. \emph{Applied Sciences} 13(3):1778.
doi:10.3390/app13031778.

Saito, T., Rehmsmeier, M. (2015). The Precision-Recall Plot Is More
Informative than the ROC Plot When Evaluating Binary Classifiers on
Imbalanced Datasets. \emph{PLOS ONE} 10(3):e0118432.
doi:10.1371/journal.pone.0118432.

Sarfraz, M. S., Chen, M.-Y., Layer, L., Peng, K., Koulakis, M. (2024).
Position: Quo Vadis, Unsupervised Time Series Anomaly Detection?
\emph{ICML}, PMLR 235:43461--43476.

Sehili, M. A., Zhang, Z. (2023). Multivariate Time Series Anomaly
Detection: Fancy Algorithms and Flawed Evaluation Methodology.
arXiv:2308.13068.

Sørbø, S., Ruocco, M. (2024). Navigating the Metric Maze: A Taxonomy of
Evaluation Metrics for Anomaly Detection in Time Series. \emph{Data
Mining and Knowledge Discovery} 38(3). doi:10.1007/s10618-023-00988-8.

Su, Y., Zhao, Y., Niu, C., Liu, R., Sun, W., Pei, D. (2019). Robust
Anomaly Detection for Multivariate Time Series through Stochastic
Recurrent Neural Network. \emph{KDD}. doi:10.1145/3292500.3330672.

Tatbul, N., Lee, T. J., Zdonik, S., Alam, M., Gottschlich, J. (2018).
Precision and Recall for Time Series. \emph{NeurIPS} 31.
arXiv:1803.03639.

Wagner, D., Michels, T., Schulz, F. C. F., Nair, A., Rudolph, M., Kloft,
M. (2023). TimeSeAD: Benchmarking Deep Multivariate Time-Series Anomaly
Detection. \emph{TMLR}.

Wagner, D., Nair, A., Franks, B. J., Arweiler, A., Muraleedharan, V. K.,
et al.~(2025). Formally Exploring Time-Series Anomaly Detection
Evaluation Metrics. arXiv:2510.17562.

Wu, R., Keogh, E. J. (2021). Current Time Series Anomaly Detection
Benchmarks are Flawed and are Creating the Illusion of Progress.
\emph{IEEE TKDE}. doi:10.1109/TKDE.2021.3112126.

Xu, H., Chen, W., Zhao, N., et al.~(2018). Unsupervised Anomaly
Detection via Variational Auto-Encoder for Seasonal KPIs in Web
Applications. \emph{WWW}, 187--196. doi:10.1145/3178876.3185996.

Yang, K., et al.~(2025). A Problem-Oriented Taxonomy of Evaluation
Metrics for Time Series Anomaly Detection. arXiv:2511.18739.

Yeh, C.-C. M., Zhu, Y., Ulanova, L., et al.~(2016). Matrix Profile I:
All Pairs Similarity Joins for Time Series. \emph{IEEE ICDM},
1317--1322. doi:10.1109/ICDM.2016.0179.

Zhong, Z., Yu, Z., Xi, X., Xu, Y., Cao, W., Yang, Y., Yang, K., You, J.
(2024). SimAD: A Simple Dissimilarity-based Approach for Time Series
Anomaly Detection. arXiv:2405.11238.

\end{document}